\documentclass[twocolumn,10pt]{IEEEtran}
\usepackage{geometry}
\newgeometry{left=0.80in,right=0.80in,top=1.0in,bottom=1.0in}
\usepackage{graphics} 
\usepackage{amsmath} 
\usepackage{amsfonts}
\usepackage{bm}
\usepackage{tikz}
\usepackage{pgfplots}
\usepackage{subcaption}

\usepackage{caption} 
\usepackage{lipsum}
\usepackage{cuted}
\usepackage{enumerate}
\usepackage{booktabs} 
\usepackage{graphicx}
\usepackage{float}
\usepackage{mathrsfs}

\usepackage{algpseudocode}
\usepackage{algorithm}
\algnewcommand\algorithmicforeach{\textbf{for each}}
\algdef{S}[FOR]{ForEach}[1]{\algorithmicforeach\ #1\ \algorithmicdo}
\usepackage{eqparbox}
\newdimen{\algindent}
\algnewcommand\LeftComment[2]{%
\hspace{#1\algindent}$\triangleright$ \eqparbox{COMMENT}{#2} \hfill %
}
\algnewcommand\LeftCommentNoTriangle[2]{%
\hspace{#1\algindent} \eqparbox{COMMENT}{#2} \hfill %
}
\begin{document}
\title{\LARGE \bf
Decentralized Multi-Agent Reinforcement Learning  for Task Offloading  Under Uncertainty\thanks{The work was supported by a Discovery Grant from the Natural Sciences and Engineering Research Council of Canada (NSERC).}
}
\author{Yuanchao Xu, Amal Feriani, and Ekram Hossain, \textit{Fellow, IEEE}}

\maketitle

\begin{abstract}
Multi-Agent Reinforcement Learning (MARL) is a challenging subarea of Reinforcement Learning due to the non-stationarity of the environments and the large dimensionality of the combined action space. Deep MARL algorithms have been applied to solve different task offloading problems. However, in real-world applications, information required by the agents (i.e. rewards and states) are subject to noise and alterations. The stability and the robustness of deep MARL to practical challenges is still an open research problem. In this work, we apply state-of-the art MARL algorithms to solve task offloading with reward uncertainty. We show that perturbations in the reward signal can induce decrease in the performance compared to learning with perfect rewards. We expect this paper to stimulate more research in studying and addressing the practical challenges of deploying deep MARL solutions in wireless communications systems.  

\let\thefootnote\relax\footnotetext{The authors are with the Department of Electrical 
and Computer Engineering, University of Manitoba, Winnipeg, MB, Canada (e-mails: xuy2@myumanitoba.ca, feriania@myumanitoba.ca, ekram.hossain@umanitoba.ca)}

\end{abstract}
\vspace{0.2cm}

\section{Introduction}
Reinforcement learning (RL) is an area which has achieved tremendous success during last few years, among all areas of Artificial Intelligence (AI). Among the recent success of reinforcement learning, most notable are, game playing at human or superhuman level~\cite{ref_minh_atari, ref_silver_go}, robotic control~\cite{ref_kober_robotics, ref_lillicrap_ddpg}, autonomous driving~\cite{ref_shalev_driving} etc. As exciting as these applications are, they cover a very small subset of the problems of practical interest, mainly because these applications fall under a category of single-agent reinforcement learning, and most practical systems have more than one agent acting together to achieve goals, which may be cooperative or in conflict with each other. This scenario falls under the category of Multi-Agent Reinforcement Learning (MARL). 

The main challenge in MARL is that, since multiple agents are learning simultaneously, the environment becomes non-stationary for each agent. Additionally, due to multiple agents acting simultaneously, the transition and reward dynamics are governed by joint action of all agents. The size of action space increases exponentially with the number of agents, as the combined action space is a Cartesian product of individual agents' action spaces. Due to these challenges, the theoretical and algorithmic understanding of MARL is still in early stages of development.

MARL problems can be solved in a centralized or decentralized manner. In the centralized technique, a central controller collects the reward information from, and dispatches the actions to individual agents. However, this approach becomes impractical as the number of agents increases. To handle large number of agents, a decentralized framework where each agent learns its own policy to maximize individual return, is more scalable. One such approach is Multi-Agent Deep Deterministic Policy Gradient (MADDPG)~\cite{ref_lowe_maddpg}, which extends the DDPG algorithm~\cite{ref_lillicrap_ddpg} to the multi-agent setting. One of the problems specific to multi-agent settings is that the agent does not have access to reward functions of other agents. The resulting model uncertainty, as faced by the agent has to be taken  into account by the solution procedure. The solution concept for such scenario is termed as robust MARL in~\cite{ref_zhang_2020} which further extends the MADDPG model into a robust-MADDPG model. In this approach, the uncertainty is modeled as the action taken by an implicit player, called as ``nature” player, which always plays against each agent.

Multiple works have focused on applying deep MARL algorithms for task offloading problems. Most of these works rely on completely decentralized methods where the agents learn task offloading policy independently. As an example, in \cite{chen2020decentralized}, each mobile device is modeled as DDPG/DQN agent observing uniquely its local information to independently learn a task offloading policy that minimizes its energy consumption and task latency. However, the independent learner training scheme is not optimal in a cooperative multi-agent systems due to the non-stationarity of the agents. Consequently, collaborative methods are more suitable. Centralized Training and Decentralized Execution (CTDE) is a well-known technique to tackle the above mentioned problems, and MADDPG is one of the state-of-the-art deep MARL algorithms using such paradigm. This motivates the application of MADDPG to solve MEC problems and specifically task offloading. For instance, \cite{lin2021maddpg} studies the joint optimization of task hierarchical offloading and resource allocation for edge systems and solves it using an algorithm based on MADDPG. Furthermore, \cite{lu2020optimization} addresses the target server selection and offloading data rate in a MEC system where users are modeled with random mobility. We refer the reader to \cite{MARLtutorialAmal} for a more detailed overview of deep MARL applications in MEC systems.

Although the previously mentioned works showcase the effectiveness of deep MARL algorithms, little attention has been paid to the practical challenges of the deployment of these methods in real-life. It is known that DRL techniques rely on a feedback from its environment to learn optimal policies. The reward signal is crucial to the convergence of DRL algorithms \cite{wang2020rlperturbatedrewards}. In real environments, rewards are subject to different kind of perturbations or noises. Consequently, it is important to study robustness and stability of deep MARL algorithms in real scenarios where reward function is noisy. 

The rest of the paper is organized as follows. The system model, assumptions, and problem formulation are presented in Section II. The basics of MARL, and the MADDPG and Robust MADDPG (RMADDPG) approaches for dynamic task offloading are presented in Section III. The numerical results are presented in Section IV before the paper is concluded in Section V.
\section{System Model, Assumptions, and Problem formulation}
In this paper, we study a multi-user MEC network encompassing $M$ mobile users, one BS, and a MEC server, operating in a time-slotted fashion with a fixed time interval $\tau$.
We consider the use case of dynamic \emph{partial} computation offloading where each mobile device $m$ has the possibility of offloading a part of its computation load to the MEC server through a wireless link. We assume that the BS is equipped with $N$ antenna elements and we denote by $\mathbf{h}_m \in \mathbb{C}^{N \times 1}$ the uplink channel between user $m$ and the BS. Thus, the received signal at the BS is as follows:
\begin{align*}
    \mathbf{y}(t) = \sum_{m=1}^M \sqrt{p_m^o(t)} \mathbf{h}_m(t) s_m(t) + \mathbf{n}(t), 
\end{align*}
where $p_m^o$ is the offloading transmission power of user $m$, $s_m$ is the transmitted symbol, and $\mathbf{n}_t \sim \mathcal{CN}(\mathbf{0}_N, \sigma^2_n\mathbf{I}_N)$ is the additive white Gaussian noise. As shown in the equation above, each user $m$ needs to decide the power level dedicated to computation offloading. 

We assume that the BS adopts the Zero-Forcing (ZF) detection algorithm given by the pseudo-inverse of the channel matrix $\mathbf{Z}(t) = \mathbf{H}^\dagger(t)= (\mathbf{H}^{H}(t)\mathbf{H}(t))^{-1}\mathbf{H}^{H}(t)$, where $\mathbf{H}(t) = [\textbf{h}_1, \dots, \mathbf{h}_M]$. 
The user $m$'s received signal is isolated by correlating the received signal $\mathbf{y}(t)$ with the combining vector $\mathbf{z}_m(t)=[\mathbf{Z}(t)]_{mm}$, the $m$-th row of the channel pseudo-inverse matrix. Consequently, the signal-to-interference-plus-noise ratio (SINR) of the $m$-th user is obtained as
\begin{align}\label{eq:sinr}
    \gamma_m(t) = \frac{p_m^o(t)}{\sigma^2_n ||\mathbf{z}_m(t)||^2}.
\end{align}

We assume that, at each time slot $t$, each mobile user $m$ receives new computation tasks following an identically distributed random process with a mean rate $\lambda_m$. The newly arrived tasks are placed into a queue $B_m(t)$ at the beginning of each time slot $t$. Each task is characterized by its input data size $d_m(t)$ and the number of CPU cycles required per bit $L_m$. In this work, we consider tasks that can be partitioned into multiple sub-tasks that can be independently computed either locally or offloaded to the MEC server. Let us denote by $d_m^l(t)$ and $d_m^o(t)$ the number of bits processed locally and the bits to be offloaded. Hence, the user's task computation queue evolves as follows:
\begin{align}\footnotesize\label{eq:queue}
\footnotesize B_m(t) =& [B_m(t-1) - d_m^l(t) - d_m^o(t)]^{+} +\\
&\lambda_m(t+1)d_m(t+1). \nonumber
\end{align}

\vspace{0.2cm}
\textbf{Local Execution}:
At each time slot $t$, the mobile device $m$ allocates $p_m^l(t)$ for local computation. Thus, the feasible CPU frequency and the number of processed bits locally are expressed as follows: 
\begin{align*}
    f_m(t) = \sqrt[3]{\frac{p_m^l(t)}{\kappa}}, \quad d_m^l(t) = \frac{\tau f_m(t)}{L_m},
\end{align*}
where $\kappa$ is the effective switched capacitance of the processor. Therefore, user $m$ will execute the tasks from the top of the queue without exceeding $d_m^l(t)$. 

\vspace{0.2cm}
\textbf{Task Offloading}:
Based on the allocated uplink transmission power $p_m^o(t)$, user $m$ can computes its possible achievable rate given by 
\begin{align*}
    R_m(t) = B \log_2(1+\gamma_m(t)),
\end{align*}
where $B$ is the transmission bandwidth and $\gamma_m(t)$ is the SINR at time slot $t$ expressed in (\ref{eq:sinr}). Hence, user $m$ can transmit $d_m^o(t)$ bits obtained by $d_m^o(t) = \tau R_m(t)$. Similar to local execution, the user will offload tasks from the top of its task queue limited by $d_m^o(t)$. 

Consequently, the objective is to learn optimal offloading policy such that the users' energy consumption is minimized while minimizing tasks completion delays. Note that, there is a trade-off between these two objectives since minimizing the task buffering delays involves allocating more power either to local computation or offloading which will increase the users' energy consumption. Therefore, the optimization problem can be expressed as follows:
\begin{align*}
    (P) : \min_{p_m^l, p_m^o} & \quad \sum_{m=1}^M \omega_m^1 (p_m^o(t) + p_m^l(t)) + \omega_m^2 B_m(t) \\
    \text{s.t.} & \quad p_m^o(t) \in [0, P_m^o], \quad m= 1,\dots,M\\
               & \quad p_m^l(t) \in [0, P_m^l], \quad m= 1,\dots,M,
\end{align*}
where $\omega_m^1, \omega_m^2$ are non-negative weights to trade-off between the two conflicting objectives, $P_m^o$ and $P_m^l$ are the maximum powers that can be allocated to task offloading and to local computation, respectively, for each user $m$. Since the problem formulation in $(P)$ involves multiple users, we adopt a multi-agent RL approach to learn optimal energy-aware dynamic task offloading strategies. Different from other works (i.e. \cite{MARLtutorialAmal}), we focus on collaborative MARL approaches where agents cooperate together to learn optimal offloading strategies.
\section{Multi-Agent Approach for Dynamic Task Offloading}

\subsection{Background}
In this paper, we focus on policy-based MARL methods and refer the reader to \cite{zhang2021MARLsurvey} for a comprehensive survey on the state-of-the art MARL methods.

Policy-based methods aim at modeling and optimizing the policy directly, rather than implicitly learning the policy by acting greedily w.r.t. learned value functions. The policy is usually represented using function approximation, parameterized by $\theta$, i.e. $\pi_{\theta}$. Given a state $s \in \mathcal{S}$, the policy $\pi_{\theta}(a|s)$ returns the probability distribution over the action space $\mathcal{A}$. Thus, the agent aims to maximize its cumulative reward or return (defined in [\ref{eq:cumul_reward}]) by performing gradient ascent in the parameter space of the approximating function:
\begin{align}\label{eq:cumul_reward}
    J(\theta) = \sum_{s \in \mathcal{S} } \rho^{\pi}(s) \sum_{a \in \mathcal{A}} \pi_{\theta}(a {\mid} s) Q^{\pi}(s,a),
\end{align}
where $\rho^{\pi}(s)$ represents the on-policy state distribution under policy $\pi$, i.e. the probability with which a state $s$ is encountered, when the agent is following the policy $\pi$.


The policy gradient theorem gives an expression for the gradient of the return $J(\theta)$, which can be evaluated using sampled trajectories:
\begin{align*}
    \nabla_{\theta}J(\theta) \propto \sum_{s \in \mathcal{S}} \rho^{\pi}(s) \sum_{a \in \mathcal{A}} Q^{\pi}(s,a) \nabla_{\theta}\pi_{\theta}(a {\mid} s).
\end{align*}
Actor-critic algorithms make use of this theorem to adjust the weights of actor/policy networks using the $Q$-values learnt by a critic network. The policy learnt by the actor is in the form of a probability distribution over the action space. Hence, this is a stochastic policy gradient algorithm. The Deterministic Policy Gradient (DPG) algorithm~\cite{ref_silver_dpg}, instead learns a deterministic policy using a deterministic policy gradient theorem, given by:
\begin{align*}
    \nabla_{\theta} J({\theta}) = \mathbb{E}_{s \sim \rho^{\mu}} \left[ \nabla_a Q^{\mu} (s, a) \nabla_{\theta} \mu_{\theta}(s) {\mid} _{a=\mu_{\theta}(s)}\right].
\end{align*}


The Deep Deterministic Policy Gradient (DDPG)~\cite{ref_lillicrap_ddpg} algorithm tries to combine the advantages of DPG with deep Q-Network (DQN) by using experience replay and target networks to stabilize the training. DDPG extends the algorithm to continuous action space which is typical in physical control applications such as robotic systems etc.

\subsection{Algorithms}
In this work, we adopt different algorithms to solve the task offloading problem as formulated in $(P)$. As mentioned above, we focus on cooperative deep MARL algorithms. In this section, we will briefly present these methods and highlight their key differences.

\subsubsection{\textbf{MADPPG}}
The applications involving interaction between multiple agents pose challenges for the single-agent algorithms, due to the emergent behavior and complexity arising from agents co-evolving together. Value-based algorithms become unstable due to the inherent non-stationarity of MARL framework, while policy gradient faces increased variance with the number of agents. Multi-Agent Deep Deterministic Policy Gradient (MADDPG)~\cite{ref_lowe_maddpg} algorithm addresses these issues by using centralized training with decentralized execution, where the critic network has access to observations and actions of all agents during training, to stabilize the training process. During inference, the agents have only local information available to them. The MADDPG algorithm extends DDPG in the following ways:
\begin{itemize}
    \item It uses centralized critic and decentralized actors during training;
    \item During training phase, agents can use estimated policies of other agents for learning;
    \item Policy ensembling, which improves the training, by reducing variance. Due to non-stationarity of the environment in MARL settings, agents can
derive a strong policy by over-fitting to the behavior of their competitors. This is undesirable as these policies may not generalize well. To avoid that, an ensemble of policies is trained, by randomly selecting one policy out of the ensemble, per episode~\cite{ref_lowe_maddpg}.
\end{itemize}

\subsubsection{\textbf{Robust MADDPG~\cite{ref_zhang_2020}}}
In MARL, the agents' behaviors, the global reward function, and the state transition dynamics are dependent on the collective action of all agents. But the individual agents do not have accurate knowledge of the actual model of the reward functions of all agents and the transition probability model. The solution obtained from the simulation without considering this uncertainty may have poor performance in practice, known as the {\em sim-to-real gap}. One approach to characterize this uncertainty is to model it as a separate agent, called as {\em nature} agent, which acts in adversarial way to all other agents. For extending the DDPG algorithm to MARL, the policy of individual agent $i$ is parameterized with a set of parameters $\theta^i$ for $i \in \mathcal{N}$. The nature’s policy is parameterized by a set of policies $\pi_{\theta^0} := \{\pi_{\theta^{0,i}}\}_{i\in\mathcal{N}}$. The joint policy is represented by $\theta = (\theta^0, \theta^1, \cdots, \theta^N )$ as the concatenation of individual agent policy parameters, along with nature's policy parameters $\theta^0 = (\theta^{0,1}, \cdots, \theta^{0, N})$. The policy gradient theorem for the robust MARL has the form:
\begin{align*}
    \nabla_{\theta^i} J^i(\theta) &= \mathbb{E}_{s\sim\rho^{s_0}_{\pi_{\theta}}, a\sim\pi_{\theta}(\cdot {\mid} s)}\left[ \nabla \log{\pi_{\theta^i}(a^i {\mid} s)} \bar{Q}^i_{\tilde{\pi}_\theta}(s {\mid} a)\right] \\
    \nabla_{\theta^{0, i}} J^i(\theta) &= \mathbb{E}_{s\sim\rho^{s_0}_{\pi_{\theta}}, a\sim\pi_{\theta}(\cdot {\mid} s)}\left[ \nabla \pi_{\theta^{0,i}}(s)[a] \right],
\end{align*}
\noindent where, $\pi_{\theta} (a {\mid} s)$ is the Cartesian product of individual agent policies. $\rho^{s_0}_{\pi_{\theta}}$ is the discounted state visitation measure, under policy $\pi_{\theta}$, with starting state $s_0$. With these, the temporal difference update rules for the actor and critic networks can be given as:
\begin{align*}
    \delta^i_t &= \pi_{\theta^{0,i}_{t}}(s_t)[a_t] + \gamma \bar{Q}_{\omega^{i}_{t}}(s_{t+1}, a_{t+1}) - \bar{Q}_{\omega^{i}_{t}}(s_{t}, a_{t}) \\
    \omega^i_{t+1} &= \omega^i_t + \alpha_t \cdot \delta^i_t \cdot \nabla \bar{Q}_{\omega^i_t}(s_t, a_t) \\
    \theta^i_{t+1} &= \theta^i_t + \beta_t \cdot \nabla \log{\pi_{\theta^i_t}}(a^i_t {\mid} s_t) \cdot \bar{Q}_{\omega^i_t}(s_t, a_t) \\
    \theta^{0,i}_{t+1} &= \theta^{0,i}_{t} - \beta_t \cdot \nabla \pi_{\theta^{0,i}_t}(s_t)[a_t].
\end{align*}

\subsubsection{MARL formulation for dynamic task offloading}
To solve the problem $(P)$ using the above algorithms, we formulate the dynamic task offloading problem as a \emph{Decentralized Partially Observable Markov Decision Process (Dec-POMDP)} \cite{MARLtutorialAmal} (Dec-POMDP) as follows:
\begin{itemize}
    \item \textbf{Observations}: This is a multi-agent scenario, each agent receives a partial observation of its environment which includes the queue length, its uplink SINR and its channel state information. Thus, for each user $m$, the state at each time slot $t$ is given by $s_m(t) = [B_m(t), \text{SINR}(t-1), \mathbf{|h_m(t)|^2}]$.
    \item\textbf{Actions}: Each mobile device $m$ decides on the power allocation for both local computation and offloading $a_m(t) = [p_m^o(t), p_m^l(t)]$. Hence, the joint action $a(t) [p_1^o(t), p_1^l(t), \dots, p_M^o(t), p_M^l(t)]$ has a space that is continuous and contains $2M$ control variables.
    \item \textbf{Rewards}: The agents' reward functions are similar to the objective function in $(P)$ where each agent strives to minimize its power consumption and queue length: $r_m(t) = - \omega_m^1 (p_m^o(t) + p_m^l(t)) - \omega_m^2 B_m(t)$.
    \item \textbf{Transition dynamics}: As has been mentioned above, the task queue for each agent evolves according to~(\ref{eq:queue}). As for the channel $\mathbf{h_m(t)}$, we assume a  Gauss-Markov model to capture the correlation between the channel samples at each time slot $t$: $ \mathbf{h}_m(t+1) = \rho_m \mathbf{h}_m(t) + \sqrt{1-\rho^2}\mathbf{e}_m(t)$, where $\rho_m$ is the normalized channel correlation coefficient and $\mathbf{e}_m(t)$ is Gaussian noise vector uncorrelated with $\mathbf{h}_m(t)$.
\end{itemize}

\section{Numerical Results}

\subsection{Simulation Setup}
In this section, we detail the experimental setup for the reported results.
In the MEC system, we assume that the agents are homogeneous which means they have the same task arrival rate and buffer size. Each agent $m$ is located at a distance $d_m$ (in meters) from the BS. The path-loss model is given by  $G(d_m) = G_0 - 10 \alpha \log_{10}(d_m/d_0)$ in dB, where $G_0$ is the  path-loss at the reference distance, $d_0$ is the reference distance and $\alpha$ is the path-loss exponent. The simulation parameters are summarized in Table~\ref{tab:params}. For all the experiments, the actor and critic are two-layer neural networks with $64$ hidden units and ReLU non-linearity. 
We use the baseline represented in \cite{chen2020decentralized} where each mobile device is equipped with a DDPG agent and the agents learn their task-offloading policies in a completely decentralized manner. The results obtained for DDPG are compared with RMADDPG algorithm~\cite{ref_zhang_2020}. The performance of the MADDPG algorithm is not reported because it shows results close to DDPG. For both cases, results are averaged across $5$ runs to avoid impact of run-to-run variations. The standard deviation over the $5$ runs is also marked on the graph. For the RMADDPG, a truncated Gaussian noise is used to model the reward uncertainty such that $\bar{R}(s,a) =\mathcal{N}_{trunc}(R(s, a);\lambda)$, where $R(s,a)$ represents the agent's true reward. Thus, the higher the value of $\lambda$, the more uncertain the rewards are.
\begin{table}[ht!]
    \centering
\caption{Simulation parameters}
    \begin{tabular}{r|l||r|l}
    \hline
    Parameter & Value & Parameter & Value\\
    \hline
         $d_1=\dots=d_m$& $100$ &$\sigma^2_n$ & $10^{-9}$ W \\
         $\alpha$& $3$ &$P_1^o=\dots=P_M^o$ & $2$ W \\
         $d_0$ & $1$ m &$P_1^l=\dots=P_M^l$ & $2$ W \\
         $G_0$ & $-30$ dB & $\rho_m$ & $0.95$ \\
         $N$& $4$ & $\kappa$ & $10^{-27}$ \\
         $L_m$ & $500$ Cycles/bit & & \\
         \hline
    \end{tabular}
    \label{tab:params}
\end{table}
\subsection{Results and Discussion}
\begin{figure*}[ht!]
     \centering
     \begin{subfigure}[t]{0.49\textwidth}
         \includegraphics[width=\linewidth]{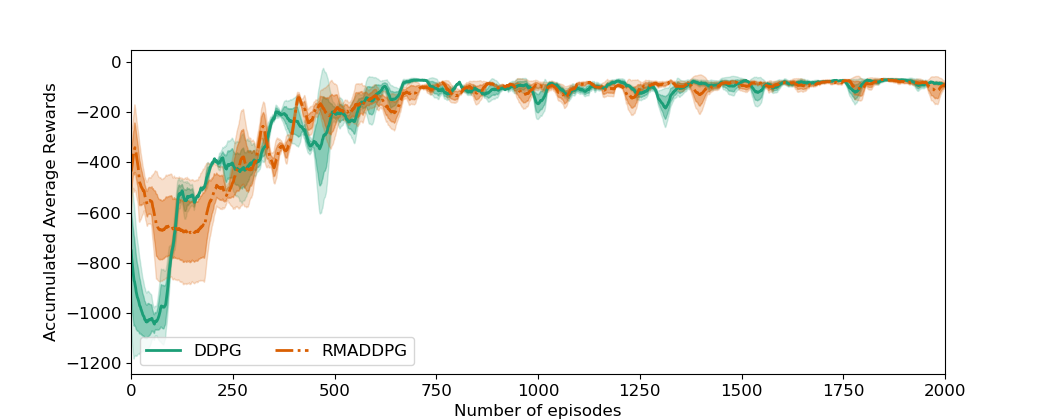}
         \caption{Perfect rewards}
         \label{fig:perfect_rew}
     \end{subfigure}
     \hfill
     \begin{subfigure}[t]{0.49\textwidth}
         \centering
         \includegraphics[width=\linewidth]{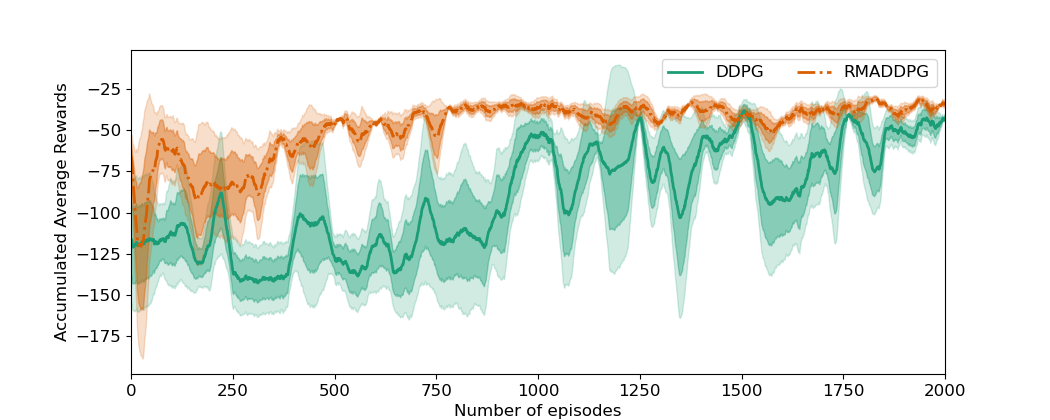}
         \caption{Noisy rewards}
         \label{fig:noise_rew}
     \end{subfigure}
        \caption{Comparison of performance under reward perturbations: (a) represents the achieved rewards during training with perfect rewards, (b) illustrates the training performance when the rewards are perturbated (noise=200).}
        \label{fig:result_1}
\end{figure*}

We start by studying the impact of the reward uncertainty on the agents training. As illustrated in Figure~\ref{fig:result_1}, it can be observed from the graph that, in presence of reward noise, the performance of both algorithms drop as compared to training without reward uncertainty. Furthermore, the agent training shows more instability when rewards are noisy. In fact,  DDPG exhibits very high variance in rewards for a large portion of the training, which can be detrimental to stability of the training process. On the contrary, the RMADPPG algorithm is more stable. In addition, it reaches convergence much faster, after around $300$ episodes, while DDPG needs close to $2000$ episodes for the training to stabilize. However, when no noise is applied, DDPG converges faster than RMADDPG and outperforms RMADDPG. Same observation is reported in \cite{ref_zhang_2020}. This shows that in ideal situations DDPG is better, but since most practical situations involve some noise in the environments, RMADDPG will perform better compared to DDPG.

\begin{figure*}[ht!]
\begin{subfigure}[t]{0.49\textwidth}
    \includegraphics[width=\linewidth]{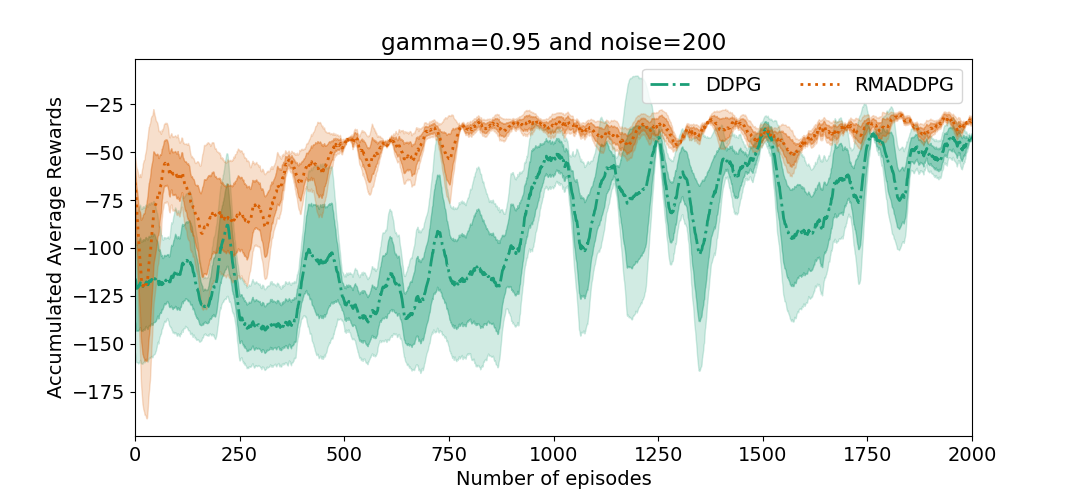}
\caption{\footnotesize$\gamma=0.95, \text{noise=200}$}
\label{fig:figure14_1}
\end{subfigure}\hfill
\begin{subfigure}[t]{0.49\textwidth}
  \includegraphics[width=\linewidth]{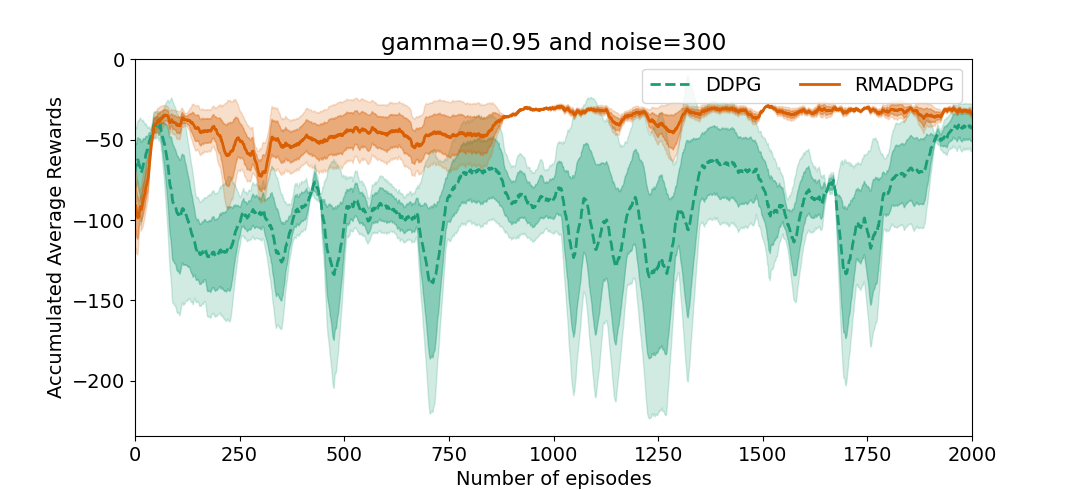}
\caption{\footnotesize$\gamma=0.95, \text{noise=300}$}
\label{fig:figure14_2}
\end{subfigure}\hfill

\begin{subfigure}[t]{0.49\textwidth}
    \includegraphics[width=\linewidth]{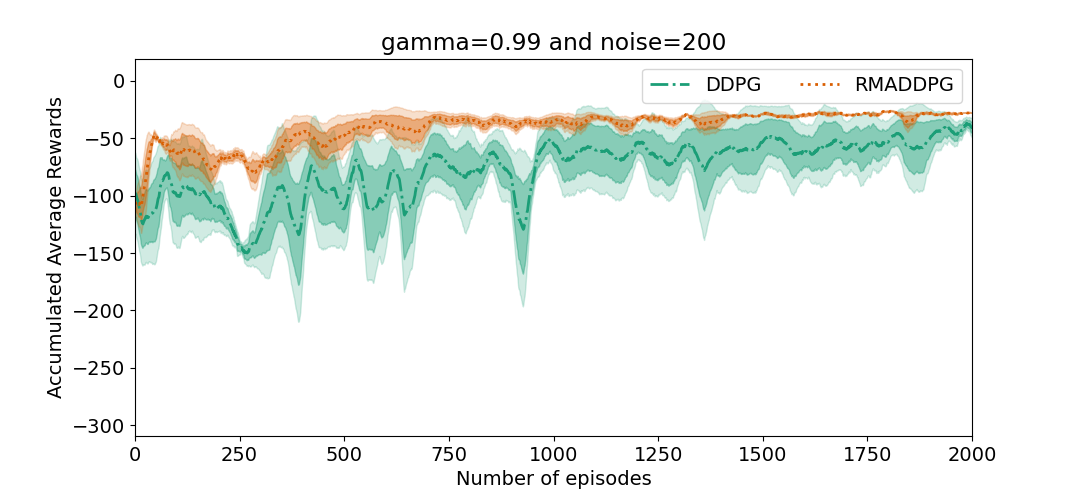}
\caption{\footnotesize$\gamma=0.99, \text{noise=200}$}
\label{fig:figure14_4}
\end{subfigure}\hfill
\begin{subfigure}[t]{0.49\textwidth}
    \includegraphics[width=\linewidth]{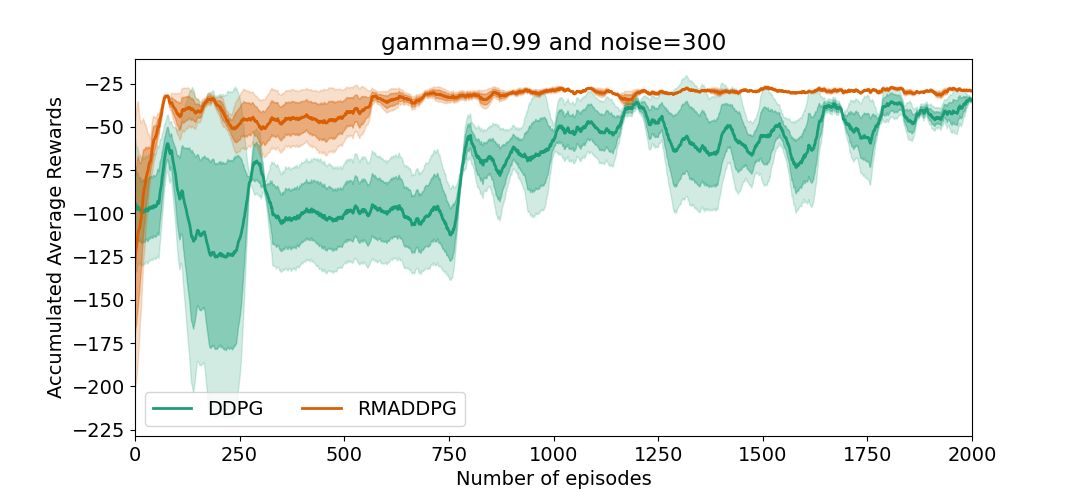}
\caption{\footnotesize$\gamma=0.99, \text{noise=300}$}
\label{fig:figure14_5}
\end{subfigure}\hfill
\caption{Cooperative task offloading: accumulated averaged rewards vs. training episodes at different reward
perturbation levels and different $\gamma$ values.}
\label{fig:results_2}
\end{figure*}

Next, we analyze the effect of $\gamma$ and $\lambda$ on the performance of both algorithms. Referring to Fig.~\ref{fig:results_2}, we can observe that the simulation reaches convergence earlier for $\gamma=0.99$ as compared to $\gamma=0.95$, for all noise levels. Higher values of $\gamma$ mean higher weight is given to future rewards. Higher values of $\gamma$ stabilize the training by reducing the reward variance.

As shown in Figure \ref{fig:results_2}, when $\lambda$ is increased, the accumulated rewards decrease. As an example, in figure \ref{fig:figure14_1}, the agents achieve a collaborative reward of $-25$ and when the reward uncertainty increases, the performance drops to $-50$ (see Figure \ref{fig:figure14_5}). For all values of $\gamma$ and reward noise, RMADDPG performs consistently better than DDPG, indicating that robustness consideration has significant impact for this case. This can also be seen in Figure~\ref{fig:results_4} which shows comparison of testing performances of both the algorithms. RMADDPG achieves higher rewards than DDPG.


\begin{figure*}[ht!]
\begin{subfigure}[t]{0.49\textwidth}
    \includegraphics[width=\linewidth]{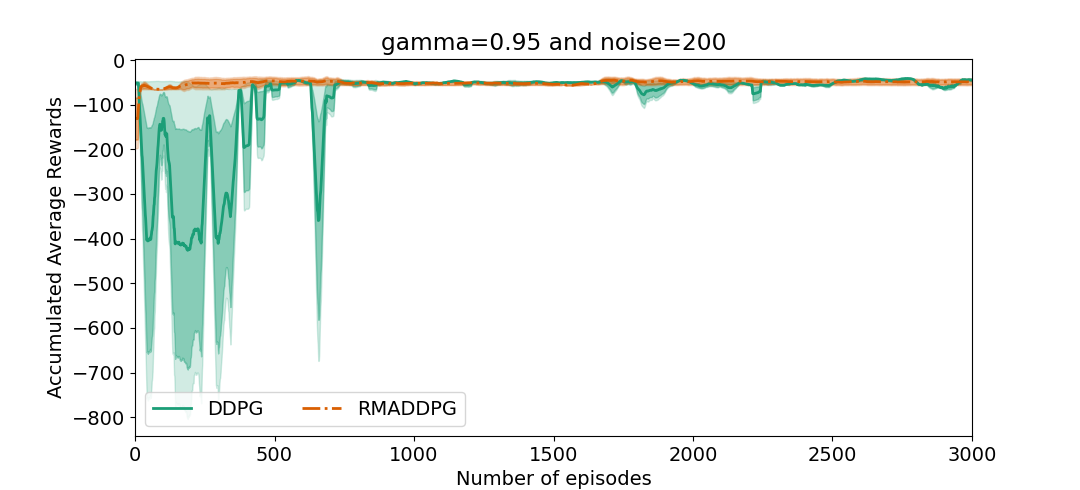}
\label{fig:figure14_7}
\end{subfigure}\hfill
\begin{subfigure}[t]{0.49\textwidth}
  \includegraphics[width=\linewidth]{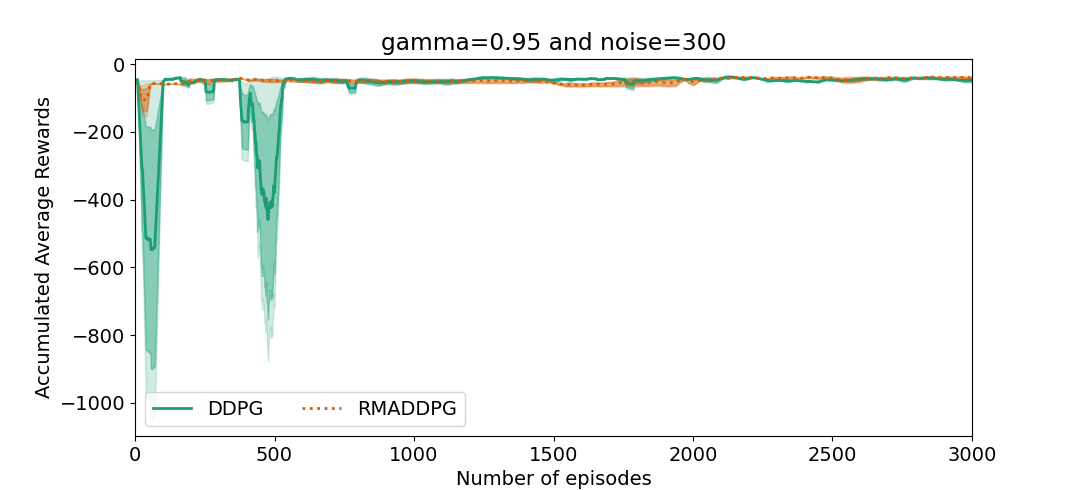}
\label{fig:figure14_8}
\end{subfigure}\hfill
\caption{DDPG vs RMADDPG for $\gamma=0.95$ and $noise=[200, 300]$ for 4 agents}
\label{fig:results_3}
\end{figure*}

\begin{figure}
    \centering
    \includegraphics[width=0.49\textwidth]{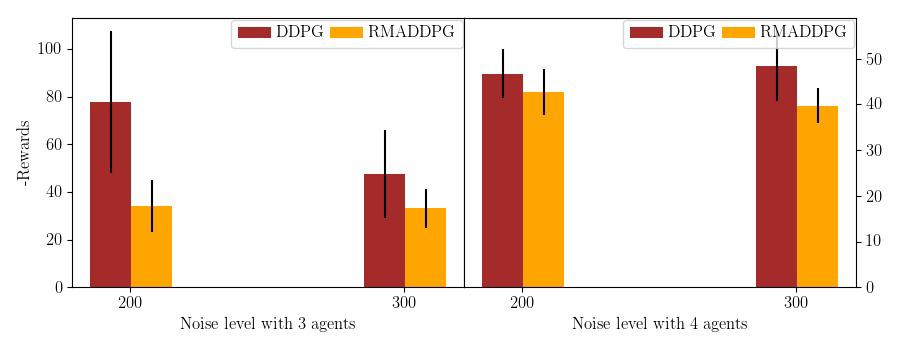}
    \caption{Test performance of DDPG vs. RMADDPG for different uncertainty levels ($\gamma=0.95$) and number of agents.}
    \label{fig:results_4}
\end{figure}

Figure~\ref{fig:results_4} shows the performance comparison for the case with $4$ agents. Here, the primary difference is that both DDPG and RMADDPG reach similar performance levels. The variance in RMADDPG is significantly smaller compared to DDPG. Also, the DDPG needs more number of episodes to reach convergence as compared to RMADDPG. Thus, it might be possible that impact of the robustness added by nature agent might get reduced as the number of agents is increased.
\section{Conclusion}
We have presented application of two actor-critic algorithms, namely, DDPG and RMADDPG, to an environment involving a continuous action space in a multi-agent setting. The task chosen consists of mobile computation offloading from mobile devices to MEC servers. We have studied the effects of system parameters such as discount rate($\gamma$), reward noise level, and number of agents. We have observed that the performance of RMADDPG  is better that DDPG across the range of parameters considered. RMADDPG exhibits a much smaller variance in average reward per step and it  converges with a much smaller number of episodes. Higher values of $\gamma$ reduce the reward variance for both algorithms. Increased reward noise adds to the variance of the average reward estimates and it makes the learning process unstable at higher values of noise levels. As the number of agents increases, the difference in performance of the two algorithms reduces and they reach the same value of maximum average reward per step.

\bibliographystyle{ieeetr}
\bibliography{refs}

\begin{thebibliography}{10}

\bibitem{ref_minh_atari}
V.~Mnih, K.~Kavukcuoglu, D.~Silver, A.~A. Rusu, J.~Veness, M.~G. Bellemare,
  A.~Graves, M.~Riedmiller, A.~K. Fidjeland, G.~Ostrovski, S.~Petersen,
  C.~Beattie, A.~Sadik, I.~Antonoglou, H.~King, D.~Kumaran, D.~Wierstra,
  S.~Legg, and D.~Hassabis, ``Human-level control through deep reinforcement
  learning,'' {\em Nature}, vol.~519, no.~7540, p.~529, 2015.

\bibitem{ref_silver_go}
D.~Silver, A.~Huang, C.~J. Maddison, A.~Guez, L.~Sifre, G.~V.~D. Driessche,
  J.~Schrittwieser, I.~Antonoglou, V.~Panneershelvam, and M.~Lanctot,
  ``Mastering the game of go with deep neural networks and tree search,'' {\em
  Nature}, vol.~529, pp.~484--489, 2016.

\bibitem{ref_kober_robotics}
J.~Kober, J.~A. Bagnell, and J.~Peters, ``Reinforcement learning in robotics: A
  survey,'' {\em The International Journal of Robotics Research}, vol.~32,
  no.~11, pp.~1238--1274, 2013.

\bibitem{ref_lillicrap_ddpg}
T.~P. Lillicrap, J.~J. Hunt, A.~Pritzel, N.~Heess, T.~Erez, Y.~Tassa,
  D.~Silver, and D.~Wierstra, ``Continuous control with deep reinforcement
  learning,'' {\em In International Conference on Learning Representations},
  2016.

\bibitem{ref_shalev_driving}
S.~Shalev-Shwartz, S.~Shammah, and A.~Shashua, ``Safe, multi-agent,
  reinforcement learning for autonomous driving,'' {\em arXiv preprint
  arXiv:1610.03295}, 2016.

\bibitem{ref_lowe_maddpg}
R.~Lowe, Y.~Wu, A.~Tamar, J.~Harb, P.~Abbeel, and I.~Mordatch, ``Multi-agent
  actor-critic for mixed cooperative-competitive environments,'' {\em In
  Advances in neural information processing systems}, pp.~6379--6390, 2017.

\bibitem{ref_zhang_2020}
K.~Zhang, T.~SUN, Y.~Tao, S.~Genc, S.~Mallya, and T.~Basar, ``Robust
  multi-agent reinforcement learning with model uncertainty,'' {\em In Advances
  in neural information processing systems}, 2020.

\bibitem{chen2020decentralized}
Z.~Chen and X.~Wang, ``Decentralized computation offloading for multi-user
  mobile edge computing: A deep reinforcement learning approach,'' {\em EURASIP
  Journal on Wireless Communications and Networking}, vol.~2020, no.~1,
  pp.~1--21, 2020.

\bibitem{lin2021maddpg}
H.~Lin, W.~Hou, H.~Wen, W.~Lei, S.~Wu, and Z.~Chen, ``Maddpg-based task
  offloading and resource management for edge system,'' in {\em The 2nd
  International Conference on Computing and Data Science}, pp.~1--5, 2021.

\bibitem{lu2020optimization}
H.~Lu, C.~Gu, F.~Luo, W.~Ding, S.~Zheng, and Y.~Shen, ``Optimization of task
  offloading strategy for mobile edge computing based on multi-agent deep
  reinforcement learning,'' {\em IEEE Access}, vol.~8, pp.~202573--202584,
  2020.

\bibitem{MARLtutorialAmal}
A.~Feriani and E.~Hossain, ``Single and multi-agent deep reinforcement learning
  for ai-enabled wireless networks: A tutorial,'' {\em IEEE Communications
  Surveys and Tutorials}, vol.~23, no.~2, pp.~1226--1252, 2021.

\bibitem{wang2020rlperturbatedrewards}
J.~Wang, Y.~Liu, and B.~Li, ``Reinforcement learning with perturbed rewards,''
  in {\em Proceedings of the AAAI Conference on Artificial Intelligence},
  vol.~34, pp.~6202--6209, 2020.

\bibitem{zhang2021MARLsurvey}
K.~Zhang, Z.~Yang, and T.~Ba{\c{s}}ar, ``Multi-agent reinforcement learning: A
  selective overview of theories and algorithms,'' {\em Handbook of
  Reinforcement Learning and Control}, pp.~321--384, 2021.

\bibitem{ref_silver_dpg}
D.~Silver, G.~Lever, N.~Heess, T.~Degris, D.~Wierstra, and M.~Riedmiller,
  ``Deterministic policy gradient algorithms,'' {\em ICML}, 2014.

\end{thebibliography}

\end{document}